\documentclass[10pt, a4paper]{article}

\usepackage{lrec-coling2024} 

\usepackage{xcolor}
\definecolor{darkgreen}{HTML}{015A00}
\definecolor{freshgreen}{HTML}{99FA67}
\definecolor{freshorange}{HTML}{FEC64C}
\definecolor{freshyellow}{HTML}{FCF171}
\definecolor{freshblue}{HTML}{ACD6F4}

\usepackage{graphicx}
\usepackage{tabularx}
\usepackage{soul}
\usepackage{algorithm}
\usepackage{algpseudocode}

\usepackage{amsmath}
\usepackage{natbib}
\usepackage{multibib}

\usepackage{hyperref}
 \definecolor{darkblue}{rgb}{0, 0, 0.5}
  \hypersetup{colorlinks=true, citecolor=darkblue, linkcolor=darkblue, urlcolor=darkblue}

\usepackage{xstring}

\usepackage{color}

\newcommand{\invisibleEndFunction}[0]{\algtext*{EndFunction}} 

\title{Decoding at the Speed of Thought:\\ Harnessing Parallel Decoding of Lexical Units for LLMs}

\name{
  \begin{tabular}{c}
    Chenxi Sun\textsuperscript{1}, Hongzhi Zhang\textsuperscript{2}, Zijia Lin\textsuperscript{2}, Jingyuan Zhang\textsuperscript{2}, Fuzheng Zhang\textsuperscript{2}, \\
    Zhongyuan Wang\textsuperscript{2}, Bin Chen\textsuperscript{2}, Chengru Song\textsuperscript{2}, Di Zhang\textsuperscript{2}, Kun Gai\textsuperscript{2}, Deyi Xiong\textsuperscript{1}\textsuperscript{*}\thanks{* Corresponding author} \\
  \end{tabular}
}

\address{
\textsuperscript{1}College of Intelligence and Computing, Tianjin University, Tianjin, China \\
\textsuperscript{2}Kuaishou Technology Inc., Beijing, China \\
         \{cxsun, dyxiong\}@tju.edu.cn\\
         \{
         \begin{tabular}{c}
         zhanghongzhi, 
linzijia, zhangjingyuan06, 
zhangfuzheng, \\
         wangzhongyuan, chenbin08, 
songchengru, 
zhangdi08, 
yuyue06
         \end{tabular}
         \}@kuaishou.com\\}

\abstract{
Large language models have demonstrated exceptional capability in natural language understanding and generation. However, their generation speed is limited by the inherently sequential nature of their decoding process, posing challenges for real-time applications. This paper introduces Lexical Unit Decoding (LUD), a novel decoding methodology implemented in a data-driven manner, accelerating the decoding process without sacrificing output quality. The core of our approach is the observation that a pre-trained language model can confidently predict multiple contiguous tokens, forming the basis for a \textit{lexical unit}, in which these contiguous tokens could be decoded in parallel. Extensive experiments validate that our method substantially reduces decoding time while maintaining generation quality, i.e., 33\% speed up on natural language generation with no quality loss, and 30\% speed up on code generation with a negligible quality loss of 3\%. Distinctively, LUD requires no auxiliary models and does not require changes to existing architectures. It can also be integrated with other decoding acceleration methods, thus achieving an even more pronounced inference efficiency boost. We posit that the foundational principles of LUD could define a new decoding paradigm for future language models, enhancing their applicability for a broader spectrum of applications. All codes are be publicly available at \href{https://github.com/tjunlp-lab/Lexical-Unit-Decoding-LUD-}{https://github.com/tjunlp-lab/Lexical-Unit-Decoding-LUD-}.
 \\ \newline  }

\begin{document}

\maketitleabstract

\section{Introduction}
The Transformer architecture \citep{NIPS2017_3f5ee243} has been crucial in recent advancements in Natural Language Processing (NLP) \cite{DBLP:journals/corr/abs-2005-14165, touvron2023llama}. Empirical evidences \citep{DBLP:journals/corr/abs-2303-08774, DBLP:journals/corr/abs-2305-10403, DBLP:journals/corr/abs-2203-15556, DBLP:conf/icml/ClarkCGMPHDHCB022, DBLP:journals/corr/abs-2001-08361, DBLP:journals/corr/abs-2102-01293} suggest a positive correlation between model size and performance, encouraging the continuous scaling of Large Language Models (LLMs). In this context, the decoder-only architecture has emerged as the de-facto standard. However, while this architecture facilitates rapid training, it still inherently predicts tokens sequentially. This is a constraint rooted in language modeling principles \cite{DBLP:journals/bstj/Shannon48, DBLP:conf/nips/BengioDV00}. This auto-regressive nature limits generation speed, posing challenges for real-time applications.

\begin{figure}[!t]
\begin{center}
\setlength{\abovecaptionskip}{-0.2cm}
\includegraphics[scale=0.099]{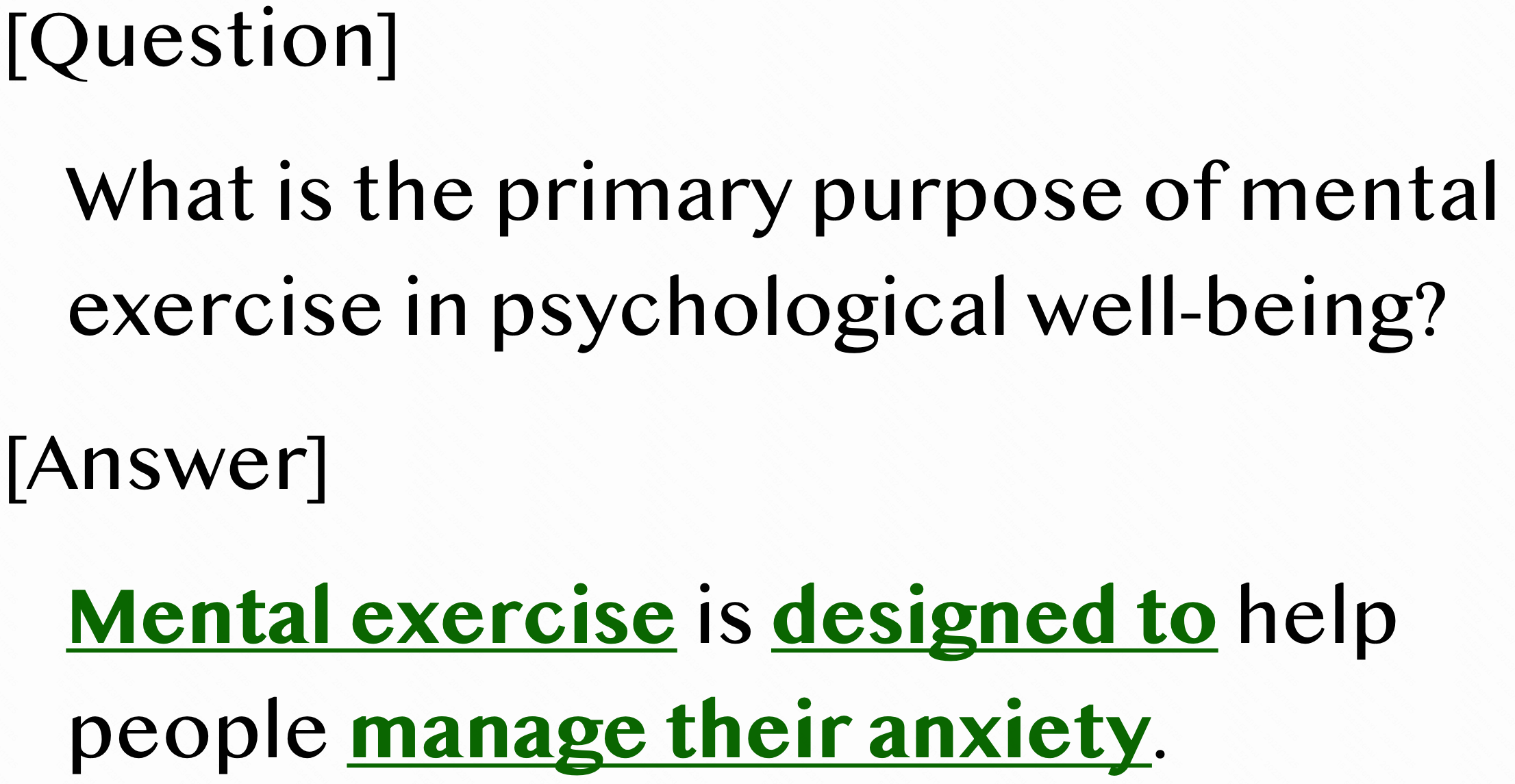} 
\caption{Illustration of ``\textcolor{darkgreen}{\underline{\textbf{lexical units}}}'' as consecutive token spans. These units, as conceptualized in our study, can potentially be identified and decoded in parallel, enhancing the decoding speed of LLMs.}
\label{fig:intuition}
\end{center}
\end{figure}

Addressing auto-regressive decoding challenges in LLMs led to numerous advancements. The initial breakthroughs occurred in machine translation with non-autoregressive transformers of encoder-decoder architectures. Those methods focus on utilizing latent variables for parallel predictions, but often sacrificed quality. Their architectural disparities generally prevent their direct applicability to accelerating LLMs \cite{DBLP:journals/corr/abs-1711-02281, DBLP:conf/icml/KaiserBRVPUS18, qian_glancing_2021, cheng_mr-p_2022, xiao_survey_2023}. Subsequent strategies have predominantly focused on computational optimization, employing techniques that reduce the complexity of models or the number of operations, though often at the expense of a certain degree of quality \cite{DBLP:journals/corr/HintonVD15, DBLP:conf/nips/JaszczurCMKGMK21, DBLP:journals/jmlr/HubaraCSEB17, DBLP:journals/corr/abs-2109-08668}.
Recent studies have revealed that some tokens are more predictable than others \cite{DBLP:journals/corr/abs-2309-02772}. Capitalizing on this insight, contemporary \textit{adaptive computation} approaches \cite{DBLP:conf/icml/LeviathanKM23} aim to efficiently predict these easier tokens and only employ complex models for challenging tokens. While those methods align with established language modeling principles and achieve desired quality levels, they often necessitate modifications to the training paradigm and the model structure \cite{DBLP:conf/acl/SchwartzSSDS20, DBLP:conf/emnlp/SchusterFJB21, medusa} or the integration of auxiliary models \cite{stern_blockwise_2018, DBLP:conf/icml/LeviathanKM23}. Such alterations can introduce additional complexities, potentially complicating the model deployment process.

In this study, we identify a notable and naturally emerging pattern within LLMs: certain span of tokens are consistently predicted with high confidence, forming what we term as ``lexical units''. The observation here intriguingly aligns with findings from linguistics and cognitive science, where humans are believed to process and produce continuous speech by segmenting it into smaller units or chunks \cite{vetchinnikova_konina_williams_mikušová_mauranen_2023}. For a visual representation of our conceptualization of lexical units, please refer to Figure~\ref{fig:intuition}.

Drawing inspiration from this observation, we introduce Lexical Unit Decoding (LUD), a novel strategy enhancing the decoding speed of LLMs.
The essence of LUD lies in the identification of 'lexical units'. A lexical unit is defined as spans of consecutive tokens predicted with high confidence by the model. This critical identification is instrumental for later fine-tuning, steering the model's capability of concurrently predicting multiple tokens during inference. LUD enables model to swiftly predict multiple tokens at once. If certainty wavers, it reverts to single-token predictions. This adaptability sets LUD apart, striking a balance between swift inference and high-quality predictions packed in one model. LUD simplifies deployment by eliminating the need for two separate models. Additionally, its compatibility with arbitrary model architectures, including the prevalent decoder-only architecture, requires no architectural modifications, further facilitating its practical application.

In our evaluations with LLaMA-13B \cite{DBLP:journals/corr/abs-2302-13971}, LUD achieves a 33\% acceleration in decoding, maintaining superb output quality. When tested on programming languages, which inherently exhibit more consistent patterns and reduced variability \cite{fu2024watermarking, kirchenbauer2023watermark}, the acceleration ratio experiences a significant upswing. This acceleration difference between natural language and code validates our method's linguistic rationality and adaptability based on content predictability. Further analysis of LUD's outputs indicates that tokens decoded concurrently by LUD invariably present coherent and linguistically meaningful units, validating our intuition that LLMs can identify these units effectively.

The elegance of our method is its deployment simplicity. Instead of resorting to complex architectural modifications, we take advantage of the model's inherent ability to generate new data based on the original dataset. The generated new data is used for continual training of parallel decoding to optimize the model's generation speed and ensure straightforward implementation. 

Our contributions can be summarized as follows.
\begin{itemize}
    \item We uncover a naturally emerging pattern within LLMs, highlighting the consistent high-confidence prediction of certain spans of tokens, which we term as ``lexical units''.
    \item We present Lexical Unit Decoding (LUD), an linguistically-adaptive, data-centric methodology that ensures lossless acceleration in decoding and seamless integration without intricate modifications of the model's architecture.
    \item We conducted an in-depth analysis on common issues in parallel decoding from a new perspective and discussed potential avenues for future research.
\end{itemize}

\section{Related Work}
\label{sec:related-work}

\begin{figure*}[!t]
\centering
\setlength{\abovecaptionskip}{0.1cm}
\includegraphics[scale=0.2]{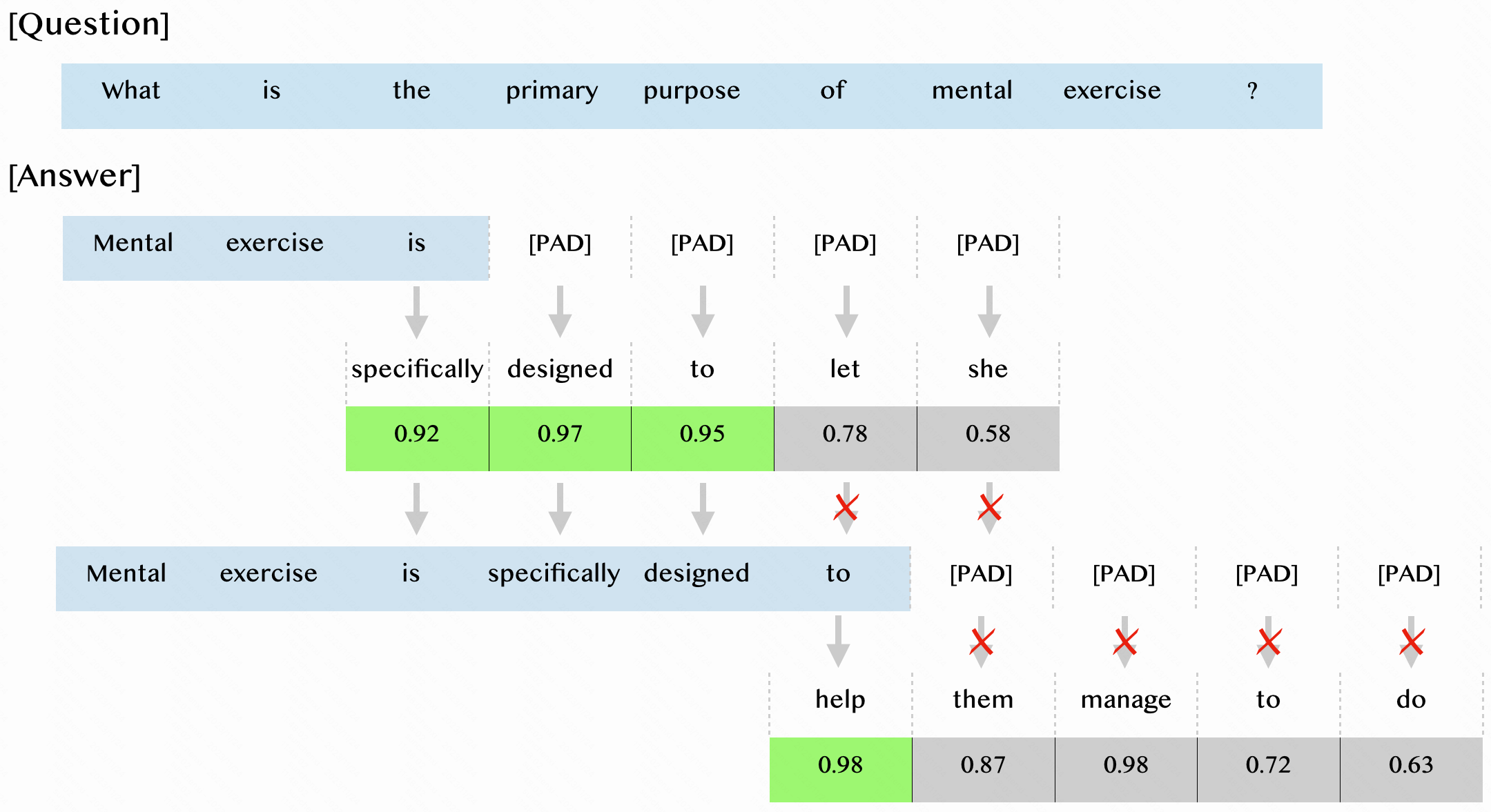} 
\caption{Illustration of the lexical unit decoding procedure. In the decoding process, we look ahead $k=5$ tokens by appending $k-1=4$ tokens and retrieving the last $k$ predicted tokens with their probabilities. However, we only accept consecutive tokens with probabilities larger than $\alpha=0.9$.
}
\label{fig:infer}
\end{figure*}

The scaling laws of LLMs have intensified the pursuit of faster inference, sparking a wave of innovation. 

The initial advancements in accelerated decoding were most prominent in machine translation, highlighted by the emergence of non-autoregressive transformers \cite{DBLP:journals/corr/abs-1711-02281}. These models, often reliant on encoder-decoder frameworks, introduced latent variables to enable parallel predictions \cite{DBLP:conf/icml/KaiserBRVPUS18, qian_glancing_2021, cheng_mr-p_2022, xiao_survey_2023}. However this approach comes at the cost of quality, thereby limiting their universality in decoder-only LLMs.

Concurrently, more general acceleration methods like distillation \cite{DBLP:journals/corr/HintonVD15}, sparcification \cite{DBLP:conf/nips/JaszczurCMKGMK21}, quantization \cite{DBLP:journals/jmlr/HubaraCSEB17}, and architectural modifications \cite{DBLP:journals/corr/abs-2109-08668} have been explored. These strategies, centered around computational optimization, seek to expedite inference with minimal performance compromise. However, they often entail an unavoidable quality reduction.

Adaptive computation has emerged as a potent strategy, with established methods like ``early exits'' leading the charge \cite{DBLP:conf/emnlp/SchusterFJB21, DBLP:journals/corr/abs-2002-07106, DBLP:conf/acl/SchwartzSSDS20, DBLP:conf/iclr/ElbayadGGA20}. Those models introduce dynamic computational depth adjustment, allowing predictions to be made earlier in the process for simpler cases, thereby enhancing inference speed \cite{DBLP:journals/cogcom/ScardapaneSBU20}.

The landscape of local-non-autoregressive adaptive computation has expanded recently. Methods aiming to decode multiple tokens simultaneously have gained traction. Innovations like separate decoding heads with tree attention mechanisms \cite{medusa}, and direct input-to-output segment copying \cite{DBLP:conf/acl/SunGWW20} demonstrate the diversity of approaches. Speculative Decoding \cite{DBLP:conf/icml/LeviathanKM23} and Blockwise Parallel Decoding \cite{stern_blockwise_2018} stand out by offering lossless acceleration, cleverly navigating the quality-speed trade-off through the use of auxiliary models. Yet, those advancements also introduce new challenges, including increased computational overhead and the complexity of integrating additional system components.

\section{Lexical Unit Decoding}

Our proposed methodology aims for efficient decoding by leveraging coherent linguistic chunks, termed as ``lexical units''. This efficiency is achieved by allowing models to predict up to $k$ continuous tokens at once as a lexical unit, rather than being confined to a single next-token prediction. The detailed approach unfolds as follows.

\subsection{Inference}

Unlike traditional models that predict one token at a time, our model, equipped with the knowledge of lexical units, attempts to predict multiple tokens in a single step, thereby accelerating the decoding process.

\paragraph{Look-Ahead Prediction}
Given a context, our model doesn't restrict itself to predicting just the immediate next token. Instead, it ambitiously casts a look-ahead window of a fixed length $k$, aiming to predict the next $k$ tokens in one sweep. This block of tokens, denoted as $x_{t:t+k}$, is predicted as:

\begin{equation}
x_{t:t+k} \sim P(x_{t:t+k} | \mathrm{context}, \mathrm{pos}(x_{t-1:t+k-1}))
\label{eq:initial_inference_block}
\end{equation}

The probability distribution of the block $x_{t:t+k}$ is conditioned on the preceding context and the positional information of the tokens before the target token within this block. In practice, we append $k-1$ \textrm{[PAD]} tokens to the context input and extract the last $k$ logits to compute the probability distribution.

\paragraph{Adaptive Span Acceptance}
While the model attempts to predict $k$ tokens, not all predictions might be of high confidence. Thus, we introduce a mechanism to selectively accept tokens from this prediction. Specifically, only the first $l$ tokens that consistently maintain probabilities above a specified threshold $\beta$ are accepted as a span $S$. $l$ is estimated as follows:
\begin{equation}
\label{eq:span_length}
l = \max(1, m)
\end{equation}
where $m$ is the largest integer such that:
\begin{align}
& \forall i \in [t, t+m]: P(x_i) \geq \beta, \nonumber \\
& P(x_{t+m+1}) < \beta. \nonumber
\end{align}

\paragraph{Token Repetition Reduction}
Parallel decoding method suffers more from the token repetition problem \cite{DBLP:journals/corr/abs-1711-02281} since the prefixing token is not ready for reference like the auto-regressive one. 
To address this issue, we implemented a straightforward method: for the decoded $k$ tokens, we check for consecutive token ids or instances where $x_{i-1}$ ends with $x_i$. If detected, the model immediately halts acceptance of new tokens.

A notable feature of our inference strategy is its adaptability. If the first token from the look-ahead window does not meet the confidence threshold, the model reverts to accepting only the first token. This ensures that in scenarios where the model isn't confident about predicting a lexical unit, it equals  
to the conventional auto-regressive decoding strategy, ensuring robustness across diverse linguistic contexts. 
The inference process is illustrated in Figure~\ref{fig:infer}.

\subsection{Data Generation}

\begin{algorithm}[!t]
\caption{Data Generation Process}
\label{alg:data_gen}
\begin{algorithmic}[1]
\Require Dataset $D$, Model $M$, Threshold $\alpha$
\Ensure Reconfigured training data $D'$
\State
\invisibleEndFunction
\Function{Identify}{$logits$, $\alpha$}
\EndFunction
\State
\Function{Reconfigure}{$item$, $lexicalUnits$}
\EndFunction
\State
\State $M_{\mathrm{FT}} \gets \text{Fine-tune } M \text{ on } D$
\State
\State $\Bar{D} \gets$ []
\For{each $item$ in $D$}
    \State $logits \gets M_{\mathrm{FT}}(item)$
    \State $lexicalUnits \gets$ \Call{Identify}{$logits$, $\alpha$}
    \State $data \gets$ \Call{Reconfigure}{$item$, $lexicalUnits$}
    \State Extend $\Bar{D}$ with $data$
\EndFor

\State $D' \gets D+\Bar{D}$

\State \Return $D'$

\end{algorithmic}
\end{algorithm}

\paragraph{Lexical Unit Identification}

Central to the methodology is the identification of ``lexical units''. These are continuous sequences of tokens that captures semantically and linguistically coherent constructs. It has been observed that pre-trained language models, especially when fine-tuned on the target dataset, tend to predict tokens within these units with remarkable confidence. 

To harness this observation, we establish an identification criterion. Specifically, a continuous span of tokens is deemed a lexical unit if the prediction probability of each token within this span surpasses a predefined threshold $\alpha$. This threshold serves as a confidence measure, ensuring that the identified spans actually represent coherent linguistic constructs.

Formally, given a span of tokens $x_{t:t+l}$, we define it as a lexical unit if the following constraints are satisfied:

\begin{enumerate}
    \item $P(x_i) \geq \alpha$ for all $i \in [t, t+l]$, where $P(x_i)$ is the prediction probability of the $i$-th token and $\alpha$ is the predefined threshold.
    \item $P(x_{t-1}) < \alpha$ and $P(x_{t+l+1}) < \alpha$, ensuring that the tokens immediately before and after the sequence have prediction probabilities lower than $\alpha$.
\end{enumerate}

To facilitate the model to recognize these lexical units, we first fine-tune the original model $M$ on the target dataset $D$. This auto-regressive training refines the model into $M_{\mathrm{FT}}$. During the forward pass with $M_{\mathrm{FT}}$ on $D$, 
by comparing probability of the ground truth label against the threshold $\alpha$, we can effectively identify spans of tokens that the model predicts with high confidence. These high-confidence spans are then designated as lexical units. This process is elaborated in Algorithm~\ref{alg:data_gen}.

\paragraph{Data Construction}

\begin{figure*}[!t]
\centering
\setlength{\abovecaptionskip}{0.1cm}
\includegraphics[scale=0.3]{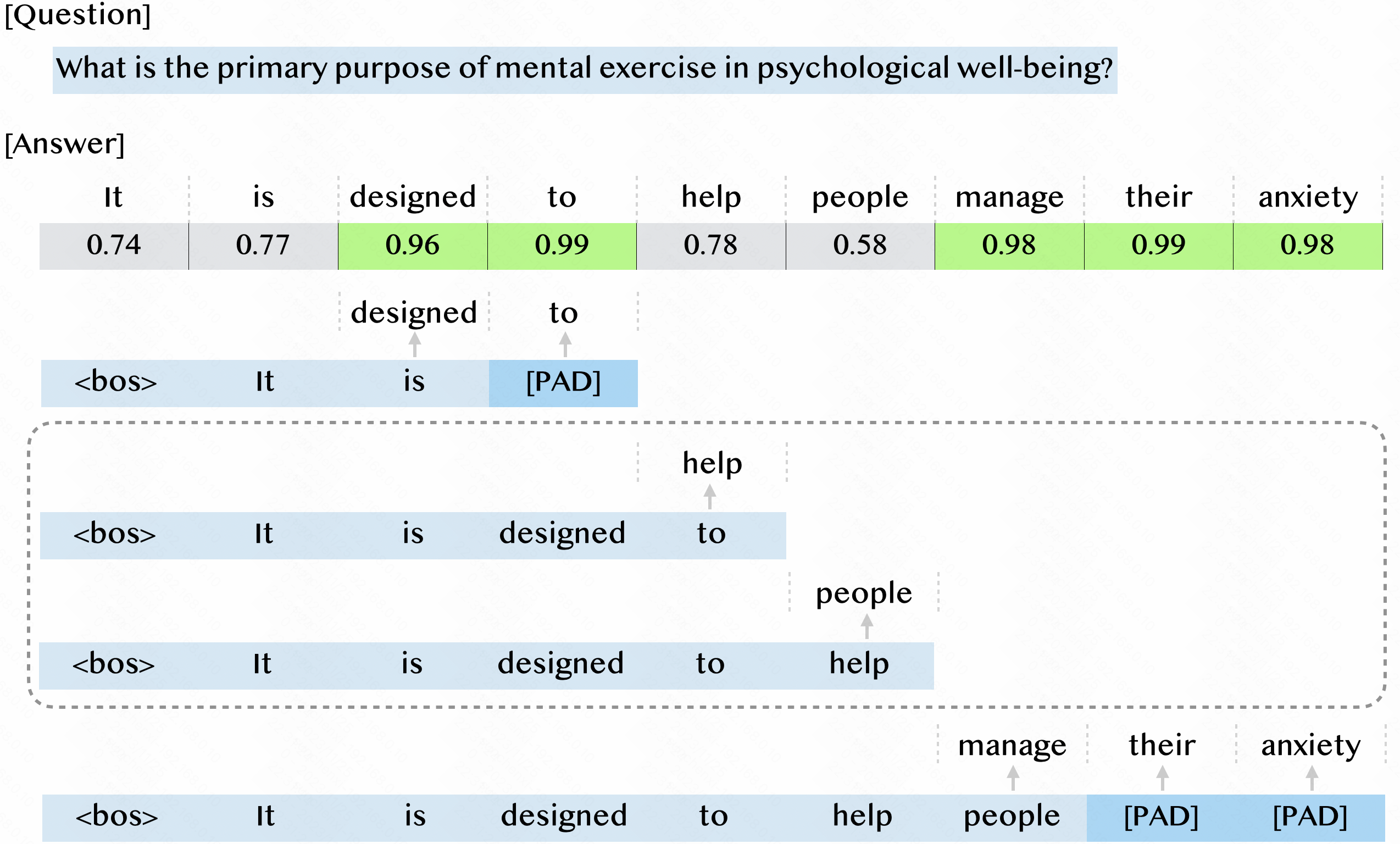} 
\caption{Visualization of the streamlined Data Generation process. Given a sequence of tokens and their corresponding probabilities, lexical units are segmented based on a threshold $\alpha=0.9$. Probabilities above the threshold are highlighted in 
green. Lexical units can consist of either a single token or multiple tokens. Multi-token lexical units are appended with \textrm{[PAD]} tokens to enable the training of parallel decoding. For individual tokens with lower prediction confidence, the model falls back to the auto-regressive training manner to maintain output quality. In practice, the second and third instances with a lexical unit of length $1$ are combined as one, which is basically the same to a standard auto-regressive training example.}
\label{fig:data_gen}
\end{figure*}

Upon identifying the lexical units, we can proceed with data reconfiguration for continual training. First, as shown in Figure~\ref{fig:data_gen}, given a single piece of data, multiple lexical units can be identified. Note that lexical units can contain multiple tokens or only a single token with coherent semantic meaning.

For every detected lexical unit, a new training instance is instantiated by replacing the tokens inside with trainable \textrm{[PAD]} tokens. This utilization of the trainable \textrm{[PAD]} token over other potential masking strategies ensures the integrity of positional information, which is critical for the model's functioning. Further more, for the new training instance, the context tokens before it are left unchanged to provide complete context information, and the loss is only calculated for the tokens within the lexical unit. In this way, we can guarantee that the loss of each token is still calculated only once even though a single sequence is split into multiple ones.

The streamlined process of data generation, from lexical unit identification to data reconfiguration, is demonstrated in Figure~\ref{fig:data_gen}. A more detailed illustration can be found in Algorithm~\ref{alg:data_gen}. Note that the generated dataset $\Bar{D}$ is mixed with the original dataset $D$ as the final reconfigured dataset $D'$.

\subsection{Training}
The core of our methodology lies in its data-centric approach. By leveraging the reconfigured dataset $D'$, we ensure that the model is proficient at both recognizing and generating lexical units during the inference phase. This is achieved without deviating from conventional training procedures, highlighting the pivotal role of our data generation process.

\paragraph{Standard Token Prediction}
Tokens that are not encapsulated within any lexical units are trained in the traditional language modeling manner. In this scenario, the model predicts the next token in the sequence based on the preceding tokens, adhering to the standard auto-regressive nature of LLMs.

\paragraph{Lexical Unit Token Prediction}
When it comes to Lexical Unit Tokens, the prediction mechanism deviates slightly. The model is trained to predict these tokens by considering not just the complete information from the tokens preceding the lexical unit, but also the positional information, which is actually embedded in the trainable \textrm{[PAD]} tokens, within the lexical unit. Note that unlike setting a look-ahead window with a fixed length $k$ during inference, the length of the identified lexical units can be varied during training. This is formulated in Equation~(\ref{eq:conditioned_token}).

\begin{equation}
x_{t} \sim P(x_{t} | x_{1:i}, \mathrm{pos}(x_{i:t-1}) )
\label{eq:conditioned_token}
\end{equation}
where $x_{t}$ denotes the Lexical Unit Tokens to be predicted and $i$ denotes the initial position of the lexical unit to which $x_{t}$ pertains. Similar to inference, its prediction is conditioned on Context Tokens \( x_{1}, x_{2}, ..., x_{i} \) preceding the lexical unit, and the positional information \( \mathrm{pos}(x_{i:t-1}) \) of the Prefixing Tokens within the lexical unit. This formulation underscores that, although the training procedure remains unchanged, the underlying training dynamics differs based on a token's association with a lexical unit.

As mentioned before, in practice, token ids of the Lexical Unit Tokens are replaced with \textrm{pad\_token\_id}, and labels for Context Tokens are set to $-100$ to exclude them from loss calculation. Note that for Lexical Unit Tokens, the model is trained to predict the original tokens rather than \textrm{[PAD]} token. \textrm{[PAD]} tokens are to provide the positional context.

\section{Experiments}

In this section, we detail the experiments conducted to evaluate the efficacy of our method. Despite the existence of numerous tasks across various domains \cite{moradshahi-etal-2023-x, liu-etal-2023-tab, ge-etal-2021-chinese}, our evaluation specifically concentrates on text and code generation tasks to thoroughly validate the adaptive acceleration capabilities of our method.

\subsection{Experimental Setup}
We used LLaMA-13B as the LLM for both experiments. Supervised Fine-tuning (SFT) is conducted with identical hyper-parameters and prompt templates as specified in Alpaca \cite{alpaca} and Code Alpaca \cite{codealpaca}. During data generation for continual training, we set $\alpha$ to $0.85$. The fine-tuned models then undergo continual training on the generated data with a batch size of $768$ and a learning rate of $3e-5$ for $3$ and $5$ epochs for text and code respectively. During inference, 
we vary $\beta$ from $0.75$ to $1.0$ while fix the size of the look-ahead window, $k$, as $10$. 

\subsubsection{Text Generation}
\paragraph{Dataset}
We used the dataset released along with Alpaca \citep{alpaca} as the training dataset and the dataset released by \citet{DBLP:conf/acl/WangKMLSKH23} as the test set. This high-quality instruction-following training dataset comprises 52,000 unique instructions generated using the self-instruction technique proposed by \citet{DBLP:conf/acl/WangKMLSKH23}. Specifically, 175 manually-written tasks are utilized as seed tasks, which serve as in-context learning examples to prompt text-davinci-003 for generating more diverse and high-quality data items. The test set comprises $252$ instructions across various domains. Following \citet{DBLP:journals/corr/abs-2304-10453}, a pairwise comparison between the generations from the fine-tuned model and its LUD version was executed. The answers from both models were fed as a pair with a prompt template to obtain two scores indicating the quality of answers respectively. To mitigate the effect of order when prompting GPT-4 for scoring, each pair was scored twice in exchanged order, and the mean score was computed as the final result.

\paragraph{Quality Metric}
The objective is to compare the quality loss before and after the LUD training process. Each scored pair can yield three possible outcomes - higher, same, or lower. We calculate the number of examples for each result and compute the quality metric as follows:

\begin{equation}
\label{eq:gsb}
R_{\mathrm{quality}} = \frac{g + s}{b + s}\text{,}
\end{equation}
where $g$, $s$ and $b$ denote the number of examples with scores higher than, same as, and lower than those generations from Alpaca, respectively.

\subsubsection{Code Generation}
\paragraph{Dataset}
For code generation, we utilized the code instruction-following dataset released with Code Alpaca \cite{codealpaca} as the training set and HumanEval \cite{DBLP:journals/corr/abs-2107-03374} for evaluation. This training dataset contains 20,000 coding instructions along with Python solutions, generated using the self-instruction technique, analogous to the Alpaca text dataset.
HumanEval \cite{DBLP:journals/corr/abs-2107-03374} encompasses 164 hand-written coding problems, each with an average of 7.7 unit tests.
\paragraph{Quality Metric}
The $\mathrm{Pass}@1$ \cite{DBLP:journals/corr/abs-2107-03374} is adopted as the basic metric to evaluate the quality of code generation. We compare LUD with the auto-regressive baseline via calculating 
\begin{equation}
R_{\mathrm{quality}} = \frac{\mathrm{Pass}@1_{\mathrm{LUD}} }{\mathrm{Pass}@1_{\mathrm{AT}}}
\end{equation}

\subsection{Acceleration Metrics}

\paragraph{Forward Compression Ratio (FCR)}
The Forward Compression Ratio (FCR) quantifies the efficiency gains achieved by our Lexical Unit Decoding (LUD) method. In a conventional auto-regressive setting, each token generated necessitates a forward calculation, making the number of tokens generated equal to the number of forward calculations. However, LUD's capability to decode multiple tokens in a single forward pass introduces a disparity between these two numbers. FCR is defined as:

\begin{equation}
{R_{\mathrm{FCR}}} = \frac{N_{\mathrm{tokens}}-{N_{\mathrm{lexical}}}}{{N_{\mathrm{tokens}}}} 
\end{equation}

A higher FCR value signifies greater computational efficiency improvement, as it indicates that fewer forward calculations are required to produce an equivalent number of tokens.

\paragraph{Wall-time Acceleration Ratio (WAR)}
While the FCR offers a theoretical perspective on efficiency, the Wall-time Acceleration Ratio (WAR) provides a more pragmatic view. It measures the acceleration in terms of actual computation time, factoring in real-world considerations including hardware efficiency, potential parallelization, and other computational overheads. WAR is given by:

\begin{equation}
{R_{\mathrm{WAR}}} = \frac{t_{\mathrm{AT}}-{t_{\mathrm{LUD}}}}{t_{\mathrm{AT}}}
\end{equation}
where, $t_{\mathrm{AT}}$ represents the average time taken to generate a single token using the auto-regressive approach, while $t_{\mathrm{LUD}}$ denotes the corresponding time for the LUD method. In practice, we measure the time consumption of solely the generation loop excluding data loading while setting batch size to $1$. The total time consumed is then divided by the number of generated tokens to get a precise estimate of the average time per token.

\subsection{Results}

\begin{table}[!t]
\centering
\setlength{\abovecaptionskip}{0.1cm}
\begin{tabular}{c|c|c|c|c} 
      \hline
      & $R_{\mathrm{quality}}$ & $R_{\mathrm{FCR}}$ & $R_{\mathrm{WAR}}$ & $\beta$ \\ 
      \hline
      Text & 100\% & 33.24\% & 33.68\% & 0.9\\
      \hline
      Code & 97\% & 29.83\% & 29.77\% & 0.99987\\ 
      \hline
\end{tabular}
\caption{Main experimental results. }
\label{tab:results}
\end{table}

The main results are shown in Table~\ref{tab:results}. LUD substantially reduces decoding time while maintaining generation quality - 33\% speed up on natural language generation with no quality loss. And a 30\% speed up on code generation is observed with a negligible quality loss of 3\%.

\begin{figure}[!t]
\begin{center}
\setlength{\abovecaptionskip}{-0.1cm}
\includegraphics[scale=0.47]{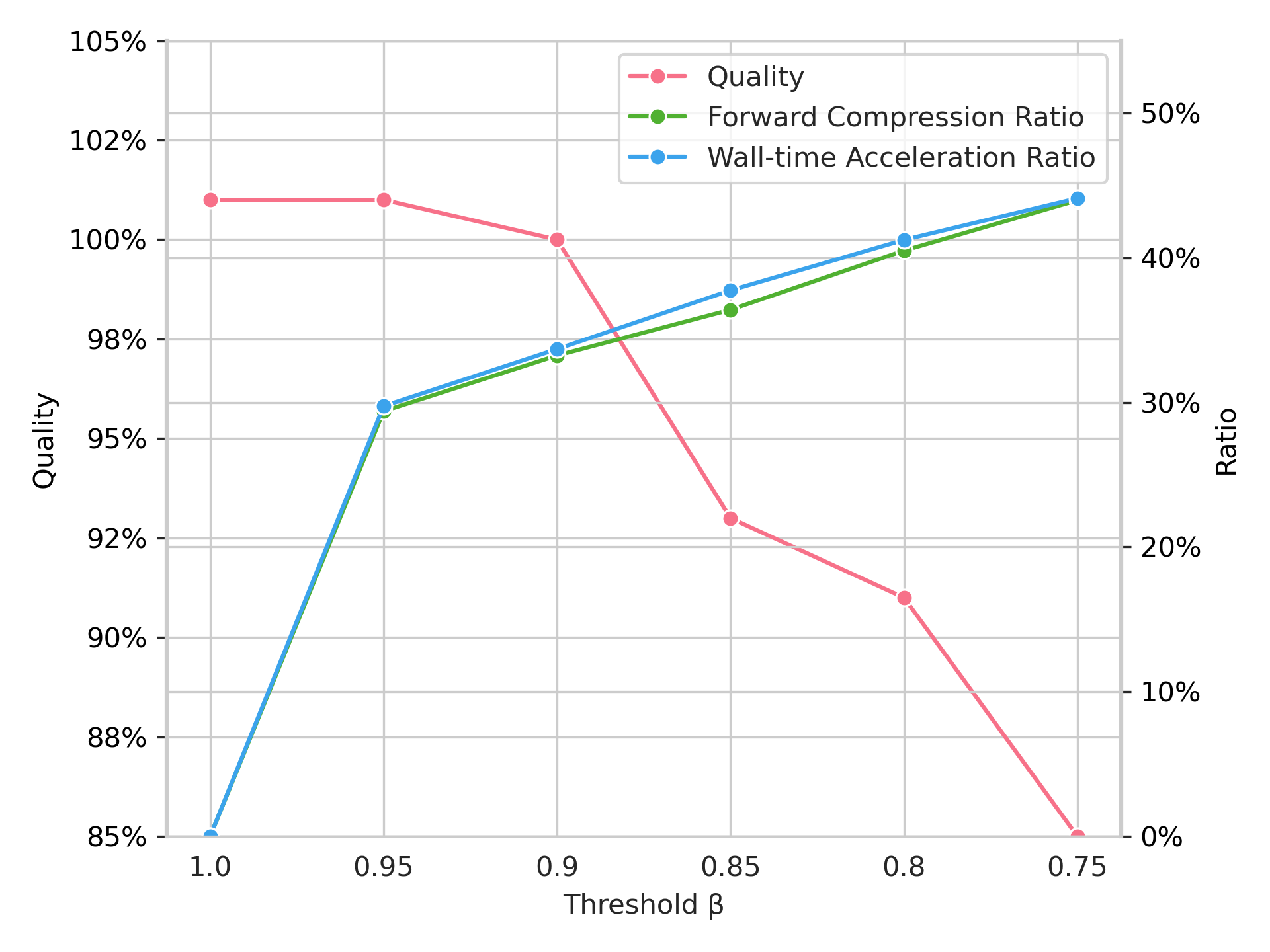} 

\caption{Quality and Acceleration curves of \textit{text} generation}

\label{fig:curves_text}
\end{center}
\end{figure}

\begin{figure}[!t]
\setlength{\abovecaptionskip}{-0.4cm}

\begin{center}

\includegraphics[scale=0.47]{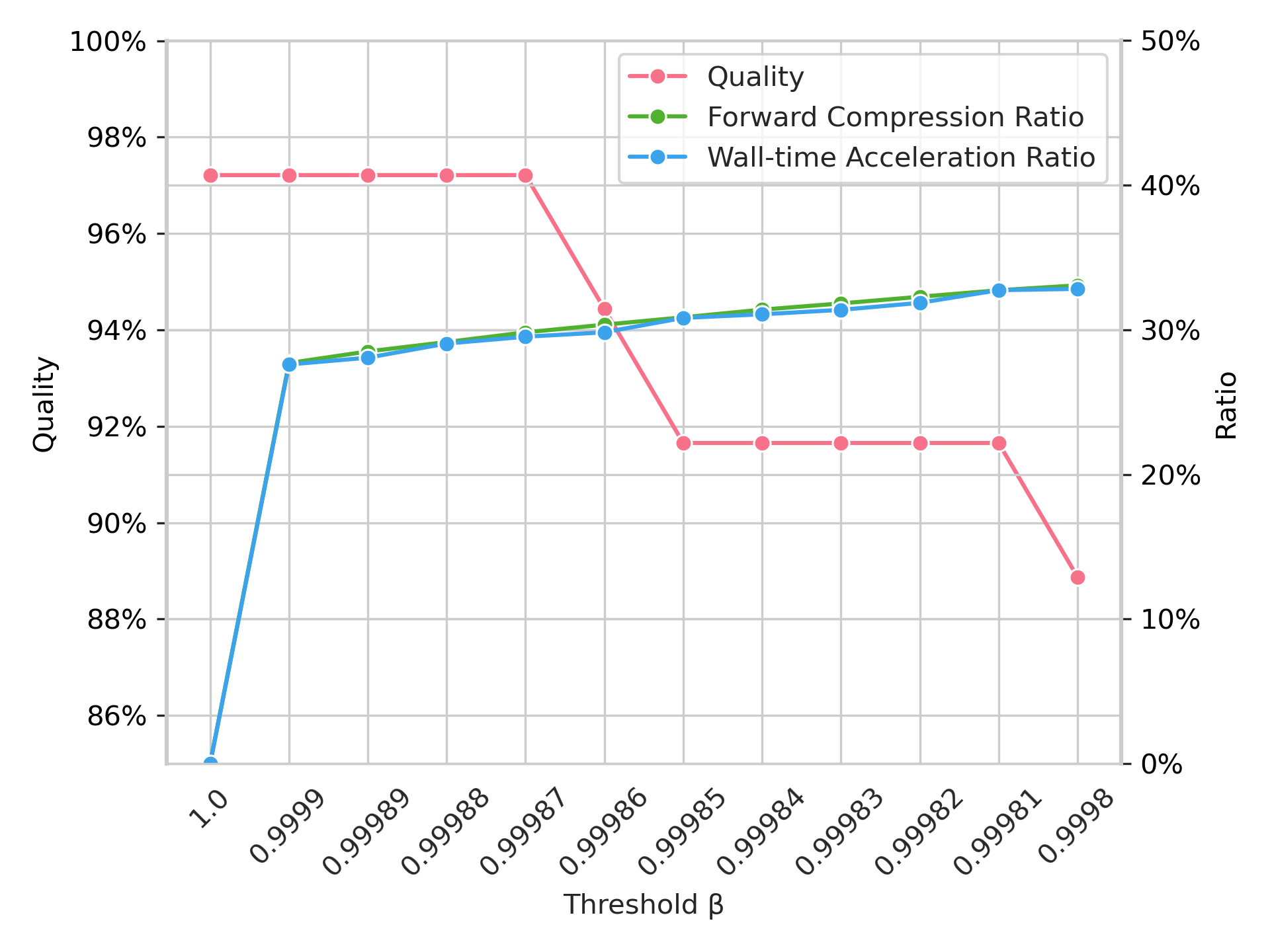} 
\caption{Quality and Acceleration curves of \textit{code} generation}
\label{fig:curves_code}
\end{center}
\end{figure}

To provide a clear and direct comparison, we also present the quality metrics alongside the acceleration curves across different $\beta$ values. These are illustrated separately for text and code in Figure~\ref{fig:curves_text} and Figure~\ref{fig:curves_code}, respectively.

\paragraph{FCR and WAR}
The efficiency of the Lexical Unit Decoding (LUD) approach is evident when examining both the FCR and WAR metrics. As we decrease the parameter \(\beta\), we observe a consistent increase in both metrics. This indicates that the model is more aggressive in accepting multiple tokens at once when the confidence threshold is lower, leading to higher acceleration. However, this efficiency achieved by a smaller $\beta$ value can cause quality degeneration.

\paragraph{Quality Loss}
The quality of the generated content, both for text and code, shows a declining trend as we push for more acceleration by decreasing \(\beta\). This degradation in quality underscores the trade-off between acceleration and quality. As we push the model to be more aggressive in its predictions, the chances of making errors increase, leading to a drop in the quality of the generated content. We do a deeper dive into the generation process in Section~\ref{section:analysis}.

\paragraph{Trade-Off and Optimal $\beta$ value}
It's apparent there's some trade-off between the acceleration ratio and quality loss when varying $\beta$. However, we did notice that the generation quality can be maintained when $\beta$ is above a specific value, within which decreasing $\beta$ can lead to faster decoding. This means there is a ``sweet pot'' of $\beta$ that achieves fastest lossless decoding. While both text and code data reconfiguration is performed with $\alpha$ set to $0.85$, the best $\beta$ values are $0.9$ and $0.99987$ respectively, which doesn't align with $\alpha$ strictly.

\paragraph{Text vs. Code Generation}
Another clear observation is with the same $\beta$ value, codes generation can be accelerated faster than text generation. The disparity between code and text generation is intriguing. One possible explanation is code sequences are viewed as a kind of low-entropy sequences while text sequences have much higher entropy \cite{kirchenbauer2023watermark}. Codes often follow specific patterns and structures, making them more predictable. This predictability might allow the model to confidently generate larger chunks of code tokens at once, leading to faster decoding acceleration. This observation validates that our method can accelerate the decoding process adaptively. Note that models for natural language generation and code generation are fine-tuned only with in-domain dataset. However, we argue that using a more balanced dataset, a sweet pot can still be achieved for a general large language model.

However, the quality of code generation degrades much faster than text as \(\beta\) decreases. This could be due to the fact that even minor errors in code can render it non-functional, whereas text might still be understandable even with minor inaccuracies. Noticeably, code decoding is also much more sensitive to $\beta$. We varied the beta value carefully and found that $0.99987$ can accelerate the decoding speed by 30\% with 3\% quality loss. We suspect that the code's attribute of being more predictable makes the model confident on its decoding. But further experiments should be conducted to understand what's happening inside the model. We leave it to our future work.

\begin{figure*}[!t]
\centering
\includegraphics[scale=0.2]{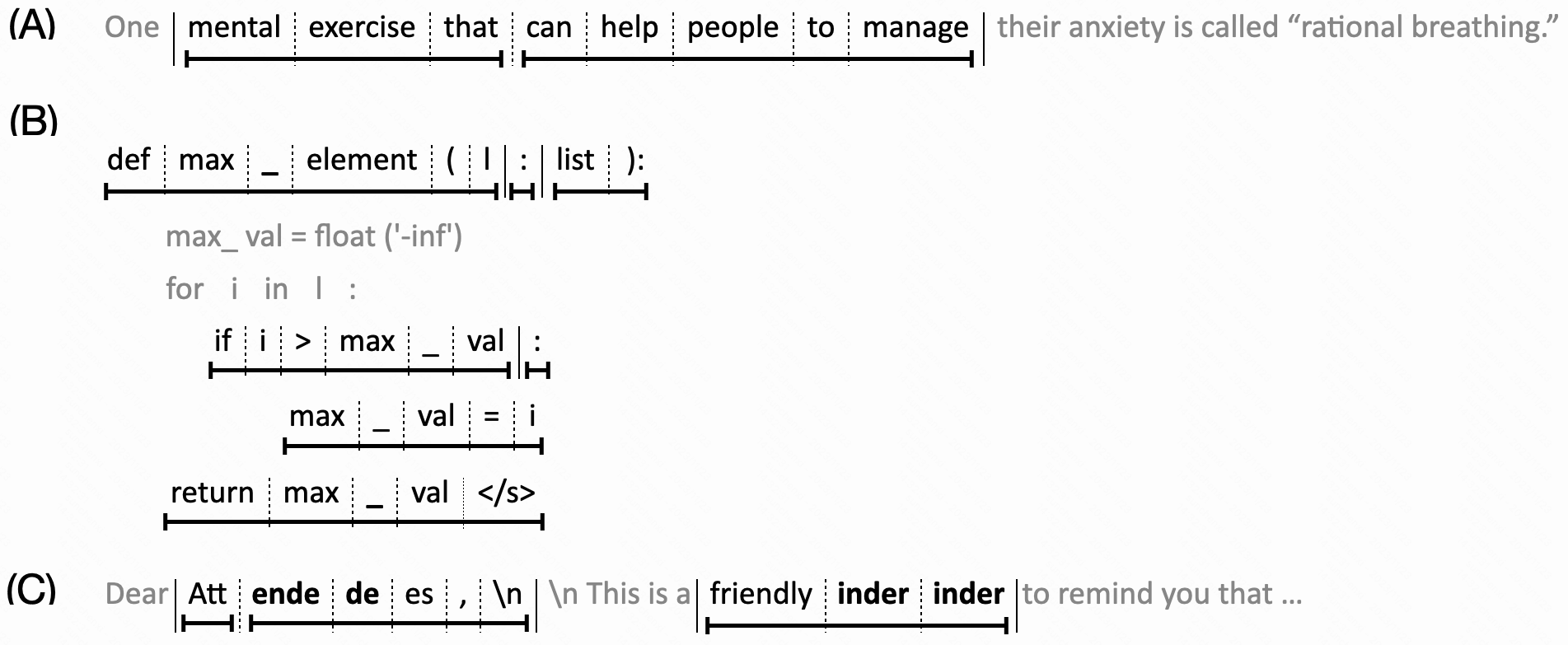} 
\caption{Examples generated with LUD. Example (A) and (B) are generated with $\beta=0.9$. Sequence (C) is generated using a smaller $\beta=0.85$ to expose the token repetition issue.}
\label{fig:examples}
\end{figure*}

\section{Analysis}
\label{section:analysis}

Our results provide a basic understanding of the Lexical Unit Decoding (LUD) approach. To further explain its behavior and implications, we delve deeper into the generation process. Given the intricate nature of those results, we perform a hands-on, empirical analysis. The codes for generating the visualization of the generation process will be publicly available.

\subsection{Coherence of Parallel Decoded Tokens}
A qualitative analysis of the tokens decoded in parallel offers a window into the model's perception of coherent linguistic constructs and grounds our definition of ``lexical units''. Our detailed inspection reveals patterns that are expected in some aspects and surprising in others. Several iconic examples are illustrated in Figure~\ref{fig:examples}. 

Example (A) showcases that LUD can indeed generate coherent constructs. We also provide example (B) as an instance of code generation. It's apparent that tokens are generated in much larger chunks leading to faster acceleration. Moreover, we have found that while tokens within a line can be highly paralleled, it's almost impossible to parallelly decode multiple tokens spanning two lines of code, except \textbackslash n and \textbackslash t. This finding echoes \citet{DBLP:journals/corr/abs-2309-02772}, where they state that the first token in a line of code is more difficult to predict than others.

\subsection{Distribution of the Number of Parallel Decoded Tokens}

\begin{figure}[!t]
\begin{center}
\setlength{\abovecaptionskip}{-0.3cm}

\includegraphics[scale=0.31]{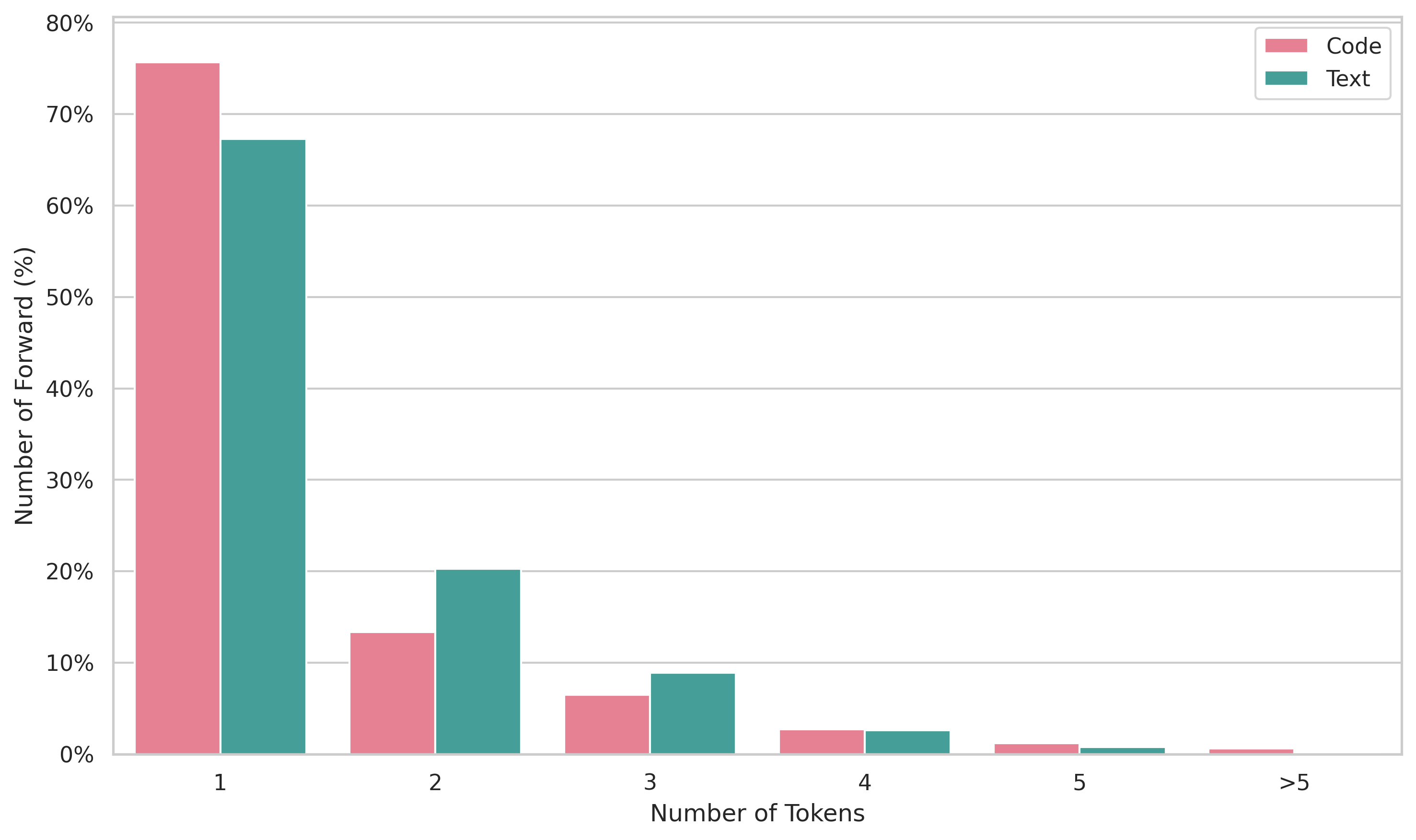} 
\caption{Distribution of the number of accepted tokens}
\label{fig:histogram}
\end{center}
\end{figure}

Figure~\ref{fig:histogram} offers an overview of the distribution of the number of accepted tokens with the window size $k=10$. The statistics is collected with $\beta=0.9$ for text and $\beta=0.99987$ for code as before.

\subsection{Token Repetition}

Parallel decoding method suffers more from the token repetition problem \cite{DBLP:journals/corr/abs-1711-02281} since the prefixing token is not ready for reference like the auto-regressive one. For consecutive token pairs $[x_{i-1}, x_i]$, we have occasionally observed token repetition during LUD's decoding process. As depicted in Figure~\ref{fig:examples} (C), repetition can manifest in various ways, such as the end of $x_i$ matching the ends of $x_{i-1}$ (partial repetition), or $x_i$ being identical to $x_{i-1}$ (identical repetition). While both repetition phenomena are rare and can be largely avoided with a larger $\beta$, empirical observation indicates that partial repetition happens much more frequent than identical repetition.

Even though, our method halts token acceptance upon detecting a repeated token, effectively reducing unnecessary repetition. However, it still permits the generation of essential repeated tokens in subsequent forward passes, recognizing when such repetitions are indeed necessary. Despite its simplicity, this approach proves effective, mitigating repetition while only marginally reducing decoding speed by 2-3\% on average.

\section{Discussion}

While our method effectively accelerate the decoding process, there are still several points worth studying: 

\paragraph{More Advanced Lexical Unit Identification Method} 
In this paper, lexical units are identified based on the prediction probability of each token. While effective, more advanced identification methods can also be further explored. For example, one can take the attention state \cite{ren-xiong-2023-huaslim} or find patterns in the neuron activation state \cite{zou2023representation} into account, then try to establish the relationship between them and meaningful lexical units, which will be used to reconfigure the dataset for parallel decoding training. Moreover, as the model undergoes continual training, its perception of lexical units might also shift so the static pre-generated data might not always be optimal. This brings forth the potential ``on-the-fly'' data generation. By dynamically generating training examples aligned with the model's current understanding, we can ensure a more harmonized training process.

\paragraph{Lexical Unit Decoding During Pre-training} 
In this work we focus on the adaption of finetuned model into parallel decoding mode via a lightweight training. It would be interesting to pre-build the lexical unit decoding capability into LLM during the pre-training procedure and we leave it as a future work.

\section{Conclusion}
In this study, we have introduced and evaluated the Lexical Unit Decoding (LUD), a novel method designed to bolster the decoding efficiency of sequence generation models. Our findings underscore the efficacy of LUD to accelerate the decoding process without sacrificing generation quality - 33\% acceleration on text generation and 30\% acceleration on code generation. We also analyze the intriguing patterns in the tokens decoded in parallel, providing insights into the model's perception of coherent linguistic constructs.

\section{Acknowledgements}
The present research was partially supported by the Key Research and Development Program of Yunnan Province (Grant No. 202203AA080004). We would like to thank the anonymous reviewers for their insightful comments.


\section{Bibliographical References}\label{sec:reference}

\bibliographystyle{lrec-coling2024-natbib}
\bibliography{lrec-coling2024-example}


\end{document}